\documentstyle[eaclap,
	tree-dvips,
	lingmacros]{article}

\input dgmacros
\input lfgmacros

\makeatletter

\newcommand{\pex}[1]{(\ref{#1})}
\newcommand{\attr}[2]{\mbox{$(#1\;#2)$}}
\newcommand{\pattr}[2]{\mbox{$\langle #1\;#2 \rangle$}}
\newcommand{\Rrel}[3]{\mbox{$R\/(#1,\;${\sc #2}$,\;#3)$}}
\newcommand{\Up}{\mbox{$\uparrow$}}

\newcommand{\Ups}{\Up_\sigma}

\newcommand{\linimp}{\;\mbox{$-\hspace*{-.4ex}\circ$}\;}
\newcommand{\All}[1]{\forall #1.\;}
\newcommand{\means}{\makebox[1.2em]{$\leadsto$}}

\newcommand{\IT}[1]{\mbox{\it #1\/}}
\newcommand{\BF}[1]{\mbox{\bf #1}}
\def\subjectarea#1{\gdef\@subject{#1}}
\def\singlespace{%
\vskip\parskip%
\vskip\baselineskip%
\def\baselinestretch{1}%
\ifx\@currsize\normalsize\@normalsize\else\@currsize\fi%
\vskip-\parskip%
\vskip-\baselineskip}
\makeatother
\begin{document}

\hyphenation{Kaplan Bresnan}

\title{\vspace{-0.5in}The Semantics of Resource Sharing in\\Lexical-Functional
Grammar}

\author{Andrew Kehler\thanks{~\tt kehler@das.harvard.edu}\\
Aiken Computation Lab\\
33 Oxford Street\\
Harvard University\\
Cambridge, MA 02138 \And
Mary Dalrymple\thanks{~\tt \{dalrymple,lamping,saraswat\}@parc.xerox.com} \\
{\bf John Lamping}\footnotemark[2] \\
{\bf Vijay Saraswat}\footnotemark[2]\\
Xerox PARC\\
3333 Coyote Hill Road\\
Palo Alto, CA 94304 }

\maketitle
\vspace{-0.5in}

\begin{abstract}
  We argue that the resource sharing that is commonly manifest
  in semantic accounts of coordination is instead appropriately
  handled in terms of structure-sharing in LFG
  \mbox{f-structures}.  We provide an extension to the previous
  account of LFG semantics \cite{DLS:EACL} according to which
  dependencies between \mbox{f-structures} are viewed as
  resources; as a result a one-to-one correspondence between uses
  of \mbox{f-structures} and meanings is maintained.  The
  resulting system is sufficiently restricted in cases where
  other approaches overgenerate; the very property of
  resource-sensitivity for which resource sharing appears to be
  problematic actually provides explanatory advantages over
  systems that more freely replicate resources during derivation.
\end{abstract}

\section{Introduction}

The resource-based approach to semantic composition in
Lexical-Functional Grammar (LFG) obtains the interpretation for a
phrase via a logical deduction, beginning with the
interpretations of its parts as premises \cite{DLS:EACL}.  The
resource-sensitive system of {\it linear logic} is used to
compute meanings in accordance with relationships manifest in LFG
\mbox{f-structures}.  The properties of the system ensure that
meanings are used exactly once, allowing {\it coherence} and {\it
completeness} conditions on \mbox{f-structures}
\cite[pages~211--212]{KaplanBresnan:LFG} to be maintained.

However, there are cases where a single constituent appears to
yield more than one contribution to the meaning of an utterance.
This is most obvious in, but is not limited to, sentences
involving coordination.  In example (\ref{ex:intro}), for
instance, {\it NAFTA} is the object of two different verbs:

\enumsentence{Bill supported, and Hillary opposed,\\ NAFTA.
\label{ex:intro} }
Since the hallmark of the linear logic approach is to ensure that
f-structure contributions are utilized exactly once in a
derivation, such constructions would at first glance appear to be
problematic for the approach.

We argue that the resource sharing that is commonly manifest in
the treatment of coordination in other approaches is
appropriately handled by exploiting the structure-sharing in LFG
\mbox{f-structures}. We refine our previous analysis to account
for cases where an \mbox{f-structure} is reached by multiple
paths from an enclosing \mbox{f-structure}.

Dalrymple et al.\ \shortcite{DLS:EACL} provides an account of LFG
semantics that represents the meaning of lexical items with
linear logic formulas.  These formulas manipulate basic
assertions of the form $f_{\sigma} \means M$, for {\em
f-structures\/} $f$ and {\em meaning logic terms\/} $M$.  Here
$\sigma$ is a mapping, the {\em semantic projection}, that
relates \mbox{f-structures} to semantic structures. To
distinguish between multiple paths entering an
\mbox{f-structure}, we now take $\sigma$ to map from sets of
paths in \mbox{f-structures} to semantic structures.  Further,
the paths between \mbox{f-structures} are made available in the
semantic space as resources.  This makes it possible for the
semantic formulas to exploit information about the multiple paths
into an \mbox{f-structure} in order to account for the multiple
uses of the \mbox{f-structure's} semantic contribution.  The
resulting system is sufficiently restricted in cases where other
approaches overgenerate; the very property of
resource-sensitivity for which resource sharing appears to be
problematic actually provides explanatory advantages over systems
that more freely replicate resources during derivation.

In Section~\ref{sec:prev}, we review previous approaches to the
semantics of coordination and argument sharing, and make note of some
of their drawbacks.  We describe the revised semantic framework in
Section~\ref{sec:LFG}, and work through several examples of
non-constituent coordination (specifically, right-node raising) in
Section~\ref{sec:examples}.  We discuss examples involving intensional
verbs in Section~\ref{sec:intensional}.

\section{Previous Work}
\label{sec:prev}

\subsection{Combinatory Categorial Grammar}

Steedman
\shortcite{Steedman:Coordination,Steedman:ConstCoord,Steedman:Gapping},
working in the framework of Combinatory Categorial Grammar (CCG),
presents what is probably the most adequate analysis of
non-constituent coordination to date.  As noted by Steedman and
discussed by Oehrle \shortcite{Oehrle:Coord}, the addition of the rule
of function composition to the inventory of syntactic rules in
Categorial Grammar enables the formation of constituents with
right-peripheral gaps, providing a basis for a clean treatment of
cases of right node raising as exemplified by sentence
(\ref{ex:intro}).  Such examples are handled by a coordination schema
which allows like categories to be conjoined, shown in
(\ref{coord-schema}).

\enumsentence{Coordination: X {\sc CONJ} X $\Rightarrow$ X
\label{coord-schema}}
This schema gives rise to various actual rules whose semantics depends
on the number of arguments that the shared material takes.  For the
cases of RNR considered here, the rule has the form shown in
(\ref{rnr-schema}).
\enumsentence{(coordination)\\[1ex]
\hspace*{-2em}X/Y:F ~{\sc CONJ}:\&~  X/Y:G  $\Rightarrow$
X/Y:{$\lambda x.(Fx \& Gx)$ \label{rnr-schema} }}
The contraction from $\lambda x.Fx$ and $\lambda x.Gx$ to $\lambda x.(Fx
\& Gx)$ in this rule allows for the single argument to be utilized
twice.

As noted by Hudson \shortcite{Hudson:RNR}, however, not all
examples of RNR involve coordinate structures:

\enumsentence{Citizens who support, paraded against politicians who
oppose, two trade bills. \label{quant-no-coord}}
Obviously, such cases fall outside of the purview of the coordination
schema.  An analysis for this sentence is available in the CCG
framework by the addition of the {\it xsubstitute} combinator
(Steedman, p.c.), as defined in Steedman
\shortcite{Steedman:NLLT}.
\enumsentence{($<$xsubstitute)\\[1ex]
Y/Z:G~(X$\backslash$Y)/Z:F $\Rightarrow$ X/Z: $\lambda x.(Fx (Gx))$}
The use of this combinator assimilates cases of noncoordinate RNR to
cases involving parasitic gaps.

While this approach has some drawbacks,\footnote{We find two problems
with the approach as it stands.  First, the intuition that one gap is
`parasitic' upon the other in cases like (\ref{quant-no-coord}) is not
strong, whereas the CCG analysis suggests an asymmetry between the two
gaps.  Second, the combinator appears to cause overgeneration.  While
it allows sentence (\ref{quant-no-coord}), it also allows sentence
(b), where {\it two trade bills} is analyzed as the object of both
verbs:
\enumsentence[(b)]{*Politicians who oppose, paraded
against, two trade bills.}} we do not offer a competing analysis of
the syntax of sentences like (\ref{quant-no-coord}) here.  Rather, we
seek an analysis of RNR (and of resource sharing in general) that is
uniform in the semantics; such a treatment isn't available in CCG
because of its tight integration between syntax and semantics.

\subsection{Partee and Rooth}
\label{sec:PR}

Perhaps the most influential and widely-adopted semantic treatment of
coordination is the approach of Partee and Rooth
\shortcite{ParteeRooth:Conjunction}.
They propose a generalized conjunction scheme in which conjuncts of
the same type can be combined.  As is the case with Steedman's
operators, contraction inherent in the schema allows for a single
shared argument to be distributed as an argument of each conjunct.
Type-lifting is allowed to produce like types when necessary; the
combination of the coordination scheme and type-lifting can have the
effect of `copying' an argument of higher type, such as a quantifier
in the case of coordinated intensional verbs.  They propose a
`processing strategy' requiring that expressions are interpreted at
the lowest possible type, with type-raising taking place only where
necessary.

To illustrate, Partee and Rooth assume that extensional verbs such as
{\it find\/} are entered in the lexicon with basic type $\langle e,
\langle e, t\rangle \rangle $, whereas intensional verbs like {\it
want\/}, which require a quantifier as an argument, have type $\langle
\langle
\langle e, t\rangle , t\rangle , \langle e, t\rangle \rangle $
(ignoring intensionality).  Two
extensional verbs such as {\it find\/} and {\it support\/} are
coordinated at their basic types:

\vbox{\enumsentence{{\it find and support\/} (type $\langle e, \langle e,
t\rangle \rangle $):

$\lambda y. \lambda x. [find(x, y) \wedge support(x, y)]$}}
\noindent Two intensional verbs such as {\it want}\/ and {\it seek}\/ are also
coordinated at their basic (higher) types:

\enumsentence{{\it want and seek} (type $\langle \langle \langle e, t\rangle ,
t\rangle , \langle e, t\rangle \rangle $):

$\lambda {\cal P}. \lambda x. [want(x, {\cal P}) \wedge seek(x, {\cal
P})]$}
The argument to this expression is a quantified NP.  When an
intensional and an extensional verb are coordinated, the extensional
verb must be type-raised to promote it to the type of the intensional
verb:

\enumsentence{{\it want and find\/} (type $\langle \langle \langle e, t\rangle
, t\rangle , \langle e, t\rangle \rangle $):

$\lambda {\cal P}. \lambda x. [want(x, {\cal P}) \wedge {\cal
P}(\lambda y. find(x, y))]$}
Again, this leads to the desired result.  However, an unwelcome
consequence of this approach, which appears to have gone unnoticed in
the literature, arises in cases in which more than two verbs are
conjoined.  If an intensional verb is coordinated with more than one
extensional verb, a copy of the quantifier will be
distributed to each verb in the coordinate structure.  For instance,
in (\ref{three-verbs}), two extensional verbs and an intensional verb
are coordinated.
\enumsentence{\label{three-verbs} {\it want, find, and support}:

$\lambda {\cal P}. \lambda x. [
\begin{array}[t]{@{\strut}l@{\strut}l@{\strut}}
& want(x, {\cal P})\\
\wedge & \hspace{.04in} {\cal P}(\lambda y.find(x, y))\\
\wedge & \hspace{.04in} {\cal P}(\lambda y.support(x, y)) \hspace{.07in} ]
\end{array}$}
Application of this expression to a quantifier results in two
quantifiers being scoped separately over the extensional verbs.  This
is the wrong result; in a sentence such as {\it Hillary wanted, found,
and supported two candidates}, the desired result is where {\it one}
quantifier scopes over both extensional verbs (that is, Hillary found and
supported the same two candidates), just as in the case where all the
verbs are extensional.  Further, there does not seem to be an obvious
way to modify the Partee and Rooth proposal so as to produce the
correct result, the problem being that the ability to copy
quantifiers inherent in their schema is too unrestricted.

A second problem with the account is that, as with Steedman's
coordination schema, Partee and Rooth's type-raising strategy only
applies to coordinate structures.  However, the need to type-raise
extends to cases not involving coordination, as in sentence
(\ref{mixed-no-coord}).

\enumsentence{\label{mixed-no-coord}
Citizens who seek, paraded against politicians who have, a decent
health insurance policy.}

We will present an analysis that preserves the intuition underlying Partee
and Rooth's processing strategy, but that predicts and generates the
correct reading for cases such as (\ref{three-verbs}).  Furthermore,
the account applies equally to examples not involving coordination, as is
the case in sentence (\ref{mixed-no-coord}).

\section{LFG and Linear Logic}
\label{sec:LFG}

LFG assumes two syntactic levels of representation: constituent
structure ({\em c-structure})\footnote{For discussion of c-structure
and its relation to \mbox{f-structure}, see, for example, Kaplan and Bresnan
\shortcite{KaplanBresnan:LFG}.} encodes phrasal dominance and
precedence relations, and functional structure ({\em f-structure})
encodes syntactic predicate-argument structure.  The \mbox{f-structure} for
sentence \pex{ex:bah} is given in \pex{ex:bahfs}:

\enumsentence{\label{ex:bah} Bill supported NAFTA.}

\enumsentence{\label{ex:bahfs} \evnup{\fd{
$f$:\fdand{\feat{\sc pred}{`{\sc support}'}
   \feat{\sc subj}{$g$:\fdand{\feat{\sc pred}{`{\sc Bill}'}}}
   \feat{\sc obj}{$h$:\fdand{\feat{\sc pred}{`{\sc nafta}'}}}}}}}
Lexical entries specify syntactic constraints on \mbox{f-structures} as well
as semantic information:

\enumsentence{\singlespace
\lexentry{Bill}{NP}{
\attr{\Up}{\sc pred} = `{\sc Bill}'\\ $\Ups \means \IT{Bill}$}

\lexentry{supported}{V}{
\attr{\Up}{\sc pred}= `{\sc support}'\\
$\All{ X, Y}\pattr{\Up}{\sc subj}_\sigma \means \mbox{X}$\\
$\qquad \otimes \hspace{.04in} \pattr{\Up}{\sc obj}_\sigma \means \mbox{Y}$ \\
$\quad \linimp \Ups \means \IT{supported}\/(X, Y)$}

\lexentry{NAFTA}{NP}{
\attr{\Up}{\sc pred} = `{\sc nafta}'\\
$\Ups \means \IT{NAFTA}$}\label{ex:bah-lex}}

\noindent Semantic information is expressed in (1) a {\em meaning
language}\/ and (2) a language for assembling meanings, or {\em glue
language}.  The meaning language could be that of any appropriate
logic; for present purposes, higher-order logic will suffice.
Expressions of the meaning language (such as {\it Bill}\/) appear on the
right side of the meaning relation \means.

The glue language is the {\em tensor\/} fragment of {\em linear
logic\/} \cite{Girard:Linear}. The semantic contribution of each
lexical entry, which we will refer to as a {\it meaning constructor},
is a linear-logic formula consisting of instructions in the glue
language for combining the meanings of the lexical entry's syntactic
arguments to obtain the meaning of the \mbox{f-structure} headed by the
entry.  For instance, the meaning constructor for the verb {\it
supported\/} is a glue language formula paraphrasable as: ``If my {\sc
subj} means $X$ and ($\otimes$) my {\sc obj} means $Y$, then
($\linimp$) my sentence means $\IT{supported}\/(X, Y)$''.

In the system described in Dalrymple et al.~\shortcite{DLS:EACL},
the \means\/ relation associates expressions in the meaning
language with \mbox{f-structures}.  As a result, each
\mbox{f-structure} contributed a single meaning constructor as a
resource to be used in a derivation.  Because linear logic does
not have any form of logical contraction (as is inherent in the
approaches discussed earlier), cases where resources are shared
appear to be problematic in this framework.  Intuitively,
however, the need for the multiple use of an \mbox{f-structure}
meaning results not from the appearance of a particular lexical
item (e.g., a conjunction) or a particular syntactic construction
(e.g., parasitic gap constructions), but instead results from
multiple paths to it from within the \mbox{f-structure} that
contains it, where structure sharing is motivated on syntactic
grounds.  We therefore revise the earlier framework to model what
we will term {\em occurrences} of \mbox{f-structures} as
resources explicitly in the logic.

\mbox{F-structures} can mathematically be regarded as (finite)
functions from a set of attributes to a set of atomic values,
semantic forms and (recursively) \mbox{f-structures}. We will
identify an occurrence of an \mbox{f-structure} with a path (from
the root) to that occurrence; sets of occurrences of an
\mbox{f-structure} can therefore be identified with path sets in
the \mbox{f-structure}. We take, then, the domain of the $\sigma$
projection to be path sets in the root \mbox{f-structure}.  Only
those path sets $S$ are considered which satisfy the property
that the extensions of each path in $S$ are identical.  Therefore
the \mbox{f-structure} reached by each of these paths is
identical. Hence from a path set $S$, we can read off an
f-structure $S_f$.  In the examples discussed in Dalrymple et
al.~\shortcite{DLS:EACL} there is a one-to-one correspondence
between the set of path sets $S$ and the set of
\mbox{f-structures} $S_f$ picked out by such path sets, so the
two methods yield the same predictions for those cases.

Relations between path sets are represented explicitly as
resources in the logic by {\it R-relations}.  \mbox{R-relations}
are represented as three-place predicates of the form
$\Rrel{F}{{\it P}}{G}$ which indicate that (the path set) $G$
appears at the end of a path $P$ (of length $1$) extending (the
path set) $F$. That is, the \mbox{f-structure } $G_f$ appears at
the end of the singleton path $P$ in the \mbox{f-structure}
$F_f$.  For example, the \mbox{f-structure} given in
(\ref{ex:bahfs}) results in two \mbox{R-relations}:

\[
\begin{array}{@{\strut}ll@{\strut}}
(i) & \Rrel{f}{subj}{g} \\
(ii) & \Rrel{f}{obj}{h}
\end{array}
\]
Because $f$ and $g$ represent path sets entering an
\mbox{f-structure} that they label, \mbox{R-relation} (i)
indicates that the set of paths \pattr{f}{\sc subj} (which
denotes the set of paths $f$ concatenated with {\sc subj}) is a
subset of the set of paths denoted by $g$.  An axiom for
interpretation provides the links between meanings of path sets
related by \mbox{R-relations}.

\[
\begin{array}{@{\strut}ll@{\strut}}
\BF{Axiom I}\colon& !  (\All{F,G,P,X} G_{\sigma} \means \IT{X} \\
& \quad \linimp !(R(F,P,G) \linimp \pattr{F}{P}_{\sigma} \means \IT{X}))
\end{array}
\]
According to this axiom, if a set of paths $G$ has meaning $X$,
then for each \mbox{R-relation} $R(F,P,G)$ that has been
introduced, a resource $\pattr{F}{P}_{\sigma} \means X$ can be
produced.  The linear logic operator `!' allows the conclusion
\mbox{$(R(F,P,G) \linimp \pattr{F}{P}_{\sigma} \means \IT{X})$}
to be used as many times as necessary: once for each
\mbox{R-relation} $R(F,P,G)$ introduced by the
\mbox{f-structure}.

We show how a deduction can be performed to derive a meaning for
example (\ref{ex:bah}) using the meaning constructors in
\pex{ex:bah-lex}, \mbox{R-relations} (i) and (ii), and Axiom I.
Instantiating the lexical entries for {\it Bill}\/, {\it NAFTA}\/, and
{\it supported}\/ according to the labels on the \mbox{f-structure} in
(\ref{ex:bahfs}), we obtain the following premises:
\[
\begin{array}{@{\strut}ll@{\strut}}
\BF{bill}\colon& g_{\sigma} \means \IT{Bill}\\
\BF{NAFTA}\colon& h_{\sigma} \means \IT{NAFTA}\\
\BF{supported}\colon& \All{ X, Y} \pattr{f}{\sc subj}_{\sigma} \means \mbox{X}
\\
& \qquad \otimes \hspace{.04in} \pattr{f}{\sc obj}_{\sigma} \means \mbox{Y} \\
 & \quad \linimp f_{\sigma} \means \IT{supported}\/(X, Y)
\end{array}
\]
First, combining Axiom I with the contribution for {\it Bill} yields:

\enumsentence{\label{ex:2a}$
\begin{array}[t]{@{\strut}l@{\strut}}
! \hspace{.03in} \forall F,P.~\Rrel{F}{{\it P}}{g} \linimp \pattr{F}{P}_\sigma
\means \IT{Bill}
\end{array}$}
This formula states that if a path set is \mbox{R-related} to the
(path set corresponding to the) \mbox{f-structure} for {\it
Bill}, then it receives {\it Bill\/} as its
meaning.  From \mbox{R-relation} (i) and formula (\ref{ex:2a}), we
derive (\ref{ex:2b}), giving the meaning of the subject of $f$.
\enumsentence{\label{ex:2b}
$\begin{array}{@{\strut}l@{\strut}}
\pattr{f}{\sc subj}_{\sigma} \means Bill
\end{array}$}
The meaning constructor for {\it supported} combines with
(\ref{ex:2b}) to derive the formula for \BF{bill-supported} shown in
(\ref{ex:2c}).
\enumsentence{\evnup{${\begin{array}{@{\strut}l@{\strut}}
\All{ Y}\pattr{f}{\sc obj}_{\sigma}\means Y\\
\qquad \linimp f_{\sigma} \means \IT{supported}\/(\IT{Bill}, Y)
\end{array}}$} \label{ex:2c}}
Similarly, using the meaning of {\it NAFTA}, R-relation (ii), and
Axiom I, we can derive the meaning shown in (\ref{ex:2cd}):
\enumsentence{$\begin{array}{@{\strut}l@{\strut}}
\pattr{f}{\sc obj}_{\sigma} \means \IT{NAFTA}
\end{array}$ \label{ex:2cd}}
and combine it with (\ref{ex:2c}) to derive
(\ref{ex:2d}):
\enumsentence{$\begin{array}{@{\strut}l@{\strut}}
f_{\sigma} \means supported(Bill, \IT{NAFTA})
\end{array}$ \label{ex:2d}}
At each step, universal instantiation and modus ponens are used.
A second derivation is also possible, in which \BF{supported}
and \BF{NAFTA} are combined first and the result is then combined with
\BF{Bill}.

The use of linear logic provides a flexible mechanism for deducing
meanings of sentences based on their \mbox{f-structure} representations.
Accounts of various linguistic phenomena have been developed within
the framework on which our extension is based, including quantifiers
and anaphora \cite{DLPS:QuantLFG}, intensional verbs
\cite{DLPS:Intensional}, and complex predicates \cite{DHLS:ROCLING}.
The logic fits well with the `resource-sensitivity' of natural
language semantics: there is a one-to-one correspondence between
\mbox{f-structure} relationships and meanings; the multiple use of resources
arises from multiple paths to them in the \mbox{f-structure}.  In the
next section, we show how this system applies to several cases of
right-node raising.

\section{Examples}
\label{sec:examples}

\subsection{RNR with Coordination}

First we consider the derivation of the basic case of right-node raising
(RNR) illustrated in sentence (\ref{ex:intro}), repeated in
(\ref{simple-coord2}).

\enumsentence{Bill supported, and Hillary opposed, \\ NAFTA.
\label{simple-coord2} }
The \mbox{f-structure} for example (\ref{simple-coord2}) is shown in
(\ref{RNR-struct}).

\enumsentence{\label{RNR-struct}\evnup{
\hspace*{-2em}\fd{$f$:
\fdset{\fdstack
{$f_{1}$:\fdand{\feat{\sc pred}{`{\sc support}'}
       \feat{\sc subj}{\node{a}{$g$:\fdand{\feat{\sc pred}{`{\sc Bill}'}}}}
       \feat{\sc obj}{\node{b}{$h$:\fdand{\feat{\sc pred}{`{\sc nafta}'}}}}}}
{$f_{2}$:\fdand{\feat{\sc pred}{`{\sc oppose}'}
       \feat{\sc subj}{\node{c}{$i$:\fdand{\feat{\sc pred}{`{\sc Hillary}'}}}}
       \feat{\sc obj}{\node{d}{\strut}}}}}}}
\nodecurve[r]{b}[r]{d}{10em}}
The meaning constructors contributed by the lexical items are as
follows:

\[\begin{array}{@{\strut}ll@{\strut}}
\BF{Bill}\colon& g_{\sigma} \means \IT{Bill}\\
\BF{Hillary}\colon& i_{\sigma} \means \IT{Hillary}\\
\BF{supported}\colon& \All{ X, Y} \pattr{f_1}{\sc subj}_{\sigma} \means X \\
& \qquad \otimes \hspace{.04in} \pattr{f_1}{\sc obj}_{\sigma} \means Y \\
& \quad \linimp f_{1 \sigma}\means \IT{supported}\/(X, Y) \\
\BF{opposed}\colon& \All{ X, Y} \pattr{f_2}{\sc subj}_{\sigma}\means X \\
& \qquad \otimes \hspace{.04in} \pattr{f_2}{\sc obj}_{\sigma} \means Y \\
& \quad \linimp f_{2 \sigma}\means \IT{opposed}\/(X, Y) \\
\BF{and}\colon& \All{ X, Y} \pattr{f}{\sc conj}_{\sigma}\means X \\
& \qquad \otimes \hspace{.04in} \pattr{f}{\sc conj}_{\sigma} \means Y \\
& \quad \linimp f_{\sigma} \means and(X, Y)\\
\BF{and2}\colon&!(\All{X,Y} \pattr{f}{\sc conj}_{\sigma} \means X \\
& \qquad \otimes  f_{\sigma} \means \IT{Y} \\
& \quad \linimp f_{\sigma} \means and(X,Y))\\
\BF{NAFTA}\colon& h_{\sigma} \means \IT{NAFTA}\\
\end{array}
\]
Here, we treat {\it and} as a binary relation.  This suffices for this
example, but in general we will have to allow for cases where more
than two constituents are conjoined.  Therefore, a second meaning
constructor \BF{and2} is also contributed by the appearance of {\it
and}, prefixed with the linear logic operator `!', so that it may be
used as many times as necessary (and possibly not at all, as is the
case in this example).

The R-relations resulting from the feature-value relationships
manifest in the f-structure in (\ref{RNR-struct}) are:\footnote{We
treat the {\sc conj} features as unordered, as they are in the
\mbox{f-structure} set.}
\[
\begin{array}{@{\strut}ll@{\strut}}
(i) & \Rrel{f}{conj}{f_1} \\
(ii) & \Rrel{f}{conj}{f_2} \\
(iii) & \Rrel{f_1}{subj}{g} \\
(iv) & \Rrel{f_1}{obj}{h} \\
(v) & \Rrel{f_2}{subj}{i} \\
(vi) & \Rrel{f_2}{obj}{h}
\end{array}
\]
There are several equivalent derivation orders; here we step through
one.\footnote{In the interest of space, we will skip some intermediate
steps in the derivation.}  Using the meanings for {\it Bill,
supported, Hillary,} and {\it opposed}, \mbox{R-relations} (iii) and (v), and
Axiom I, we can derive meanings for {\it Bill supported} and
{\it Hillary opposed} in the fashion described in Section~\ref{sec:LFG}:

\[\begin{array}[t]{@{\strut}l@{\strut}l@{\strut}}
\BF{bill-supported}\colon&
\All{ Y}\pattr{f_1}{\sc obj}_{\sigma}\means Y \\
& \quad \linimp f_{1 \sigma} \means
\IT{supported}\/(\IT{Bill}, Y) \\
\BF{hillary-opposed}\colon&
\All{ Z}\pattr{f_2}{\sc obj}_{\sigma}\means Z \\
& \quad \linimp f_{2 \sigma} \means
\IT{opposed}\/(\IT{Hillary}, Z)
\end{array}
\]
We combine the antecedents and consequents of the foregoing formulae to yield:

\[\begin{array}[t]{@{\strut}l@{\strut}}
\BF{bill-supported $\otimes$ hillary-opposed}\colon \\
\All{ Y, Z} \pattr{f_1}{\sc obj}_\sigma \means Y \otimes
\hspace{.02in} \pattr{f_2}{\sc obj}_{\sigma}\means Z \\
\quad \linimp f_{1 \sigma} \means \IT{supported}\/(\IT{Bill}, Y) \\
\qquad \otimes f_{2 \sigma} \means \IT{opposed}\/(\IT{Hillary}, Z)
\end{array}
\]
Consuming the meaning of {\it and} and \mbox{R-relations} (i) and (ii), and
using Axiom I, we derive:

\[\begin{array}[t]{@{\strut}l@{\strut}}
\BF{bill-supported-and-hillary-opposed1}\colon \\
\All{ Y, Z} \pattr{f_1}{\sc obj}_{\sigma}\means Y
\otimes \pattr{f_2}{\sc obj}_{\sigma}\means Z \\
\quad \linimp f_{\sigma} \means
and(\begin{array}[t]{@{\strut}l@{\strut}}
\IT{supported}\/(\IT{Bill}, Y),\\ \IT{opposed}\/(\IT{Hillary}, Z))
\end{array}\end{array}
\]
Using Axiom I and  \mbox{R-relations} (iv) and (vi), the following
implication can be derived:

\[\begin{array}[t]{l}
\All{ X} h_{\sigma}\means X \\
\quad \linimp \pattr{f_1}{\sc obj}_{\sigma}\means X
\otimes \pattr{f_2}{\sc obj}_{\sigma}\means X
\end{array}
\]
Using these last two formulae, by transitivity we obtain:

\[\begin{array}[t]{l}
\BF{bill-supported-and-hillary-opposed2}\colon \\
\All{ X} h_{\sigma}\means X\\
\quad \linimp f_{\sigma} \means
and(\begin{array}[t]{@{\strut}l@{\strut}}
\IT{supported}\/(\IT{Bill}, X),\\
\IT{opposed}\/(\IT{Hillary}, X))\end{array}
\end{array}
\]
Finally, consuming the contribution of {\it NAFTA}, by universal
instantiation and modus
ponens we obtain a meaning for the whole sentence:

\[\begin{array}[t]{l}
f_{\sigma} \means
and(\begin{array}[t]{@{\strut}l@{\strut}}
\IT{supported}\/(\IT{Bill},NAFTA),\\
\IT{opposed}\/(\IT{Hillary}, \IT{NAFTA}))\end{array}
\end{array}
\]
At this stage, all accountable resources have been consumed, and
the deduction is complete.

\begin{figure*}
\[\begin{array}{@{\strut}l@{\strut}l@{\strut}}
\BF{Hillary}\colon& g_{\sigma} \means \IT{Hillary}\\
\BF{wanted}\colon& \All{ X,Y} \pattr{f_1}{\sc subj}_{\sigma} \means X
\\ & \hspace*{.4in} \otimes \hspace{.03in} (\All{ s,p} (\All{X} \pattr{f_1}{\sc
subj}_{\sigma}\means X
\linimp  s \means p(X)) \linimp s \means Y(\hat{\ }{p})) \\
&  \quad \linimp  f_{1 \sigma}\means \IT{wanted}(X, \hat{\ }{Y}) \\
\BF{found}\colon& \All{ X, Y} \pattr{f_2}{\sc subj}_{\sigma}\means X
 \otimes \pattr{f_2}{\sc obj}_{\sigma} \means Y
 \linimp f_{2 \sigma}\means \IT{found}(X, Y) \\
\BF{supported}\colon& \All{ X, Y} \pattr{f_3}{\sc subj}_{\sigma} \means X
\otimes\pattr{f_3}{\sc obj}_{\sigma} \means Y
\linimp  f_{3 \sigma}\means \IT{supported}\/(X, Y) \\
\BF{and}\colon& \All{ X, Y} \pattr{f}{\sc conj}_{\sigma}\means X
\otimes \pattr{f}{\sc conj}_{\sigma} \means Y
\linimp f_{\sigma} \means and(X, Y)\\
\BF{and2}\colon &
\hspace{.01in} ! (\All{X,Y} \pattr{f}{\sc conj}_{\sigma} \means X \otimes
f_{\sigma} \means \IT{Y}
 \linimp f_{\sigma} \means and(X,Y))\\
\BF{two-candidates}\colon & \All{ H,S} (\All{ x} h_{\sigma}\means X \linimp  H
\means S(x))
\linimp H \means two\/(z, candidate(z), S(z))
\end{array}
\]\caption{Meaning constructors for sentence
(\ref{mixed-int-ent})}\label{fig:1}
\end{figure*}

\subsection{RNR with Coordination and Quantified NPs}

We now consider sentence (\ref{quant-coord}), where a quantified NP is shared.

\enumsentence{Bill supported, and Hillary opposed, two trade bills.
\label{quant-coord} }

\noindent Partee and Rooth \shortcite{ParteeRooth:Conjunction} observe,
and we agree, that the quantifier in such cases only scopes once,
resulting in the reading where Bill supported and Hillary opposed the
same two bills.\footnote{We therefore disagree with Hendricks
\shortcite{Hendriks:Flexibility}, who claims that such sentences
readily allow a reading involving four trade bills.}  Our analysis
predicts this fact in the same way as Partee and Rooth's analysis
does.

The meanings contributed by the lexical items and \mbox{f-structure}
dependencies are the same as in the previous example, except for that
of the object NP.  Following Dalrymple et
al.~\shortcite{DLPS:QuantLFG}, the meaning derived using the
contributions from an \mbox{f-structure} $h$ for {\it two trade bills}
is:

\[\begin{array}[t]{l}
\BF{two-trade-bills}\colon \\
\All{ H,S} (\All{ x} h_{\sigma}\means x \linimp  H \means S(x)) \\
\quad \linimp H \means two\/(z, trade\-bill(z), S(z))
\end{array}
\]
The derivation is just as before, up until the final step, where we
have derived the formula labeled
\BF{bill-supported-and-hillary-opposed2}.  This formula matches the
antecedent of the quantified NP meaning, so by universal instantiation
and modus ponens we derive:
\[\begin{array}[t]{@{\strut}l@{\strut}}
f_{\sigma} \means two\/(z,trade\-bill(z),
and(\begin{array}[t]{@{\strut}l@{\strut}}
\IT{supported}\/(\IT{Bill}, z),\\ \IT{opposed}\/(\IT{Hillary}, z)))
\end{array}
\end{array}
\]
With this derivation, there is only one quantifier meaning which
scopes over the meaning of the coordinated material.  A result where
the quantifier meaning appears twice, scoping over each conjunct
separately, is not available with the rules we have given thus far; we
return to this point in Section \ref{sec:intensional}.

The analysis readily extends to cases of noncoordinate RNR such as
example (\ref{quant-no-coord}), repeated as example
(\ref{q-n-c-again}).
\enumsentence{Citizens who support, paraded against politicians who
oppose, two trade bills. \label{q-n-c-again}}
In our analysis, the \mbox{f-structure} for {\it two trade bills\/} is
resource-shared as the object of the two verbs, just as it is in the
coordinated case.

Space limitations preclude our going through the derivation; however,
it is straightforward given the semantic contributions of the lexical
items and \mbox{R-relations}.  The fact that there is no coordination
involved has no bearing on the result, since the semantics of
resource-sharing is distinct from that of coordination in our
analysis.  As previously noted, this separation is not possible in
CCG because of the tight integration between syntax
and semantics.  In LFG, the syntax/semantics interface is more loosely
coupled, affording the flexibility to handle coordinated and
non-coordinated cases of RNR uniformly in the semantics.  This also
allows for our semantics of coordination not to require schemas nor
entities of polymorphic type; our meaning of {\it and} is type $t
\times t \rightarrow t$.

\section{Intensional Verbs}
\label{sec:intensional}

We now return to consider cases involving intensional verbs.
The preferred reading for sentence (\ref{mixed-int-ent}), in which
only one quantifier scopes over the two extensional predicates, is
shown below:

\enumsentence{Hillary wanted, found, and supported two
candidates. \label{mixed-int-ent}}

\[\begin{array}[t]{@{\strut}l@{\strut}}
and(wanted\/(\begin{array}[t]{@{\strut}l@{\strut}}\IT{Hillary}, \\
\hat{\ } \lambda Q.two\/(x, candidate(x), [\check{\ }{Q}](x))),\end{array} \\
\hspace*{.65in} two\/(z,\begin{array}[t]{@{\strut}l@{\strut}}
candidate(z),\\
and(\begin{array}[t]{@{\strut}l@{\strut}}
\IT{found}(\IT{Hillary}, z),\\ \IT{supported}\/(\IT{Hillary}, z))))
\end{array}\end{array}\end{array}\]

\noindent The \mbox{f-structure} for example (\ref{mixed-int-ent}) is given in
(\ref{int-struct}).

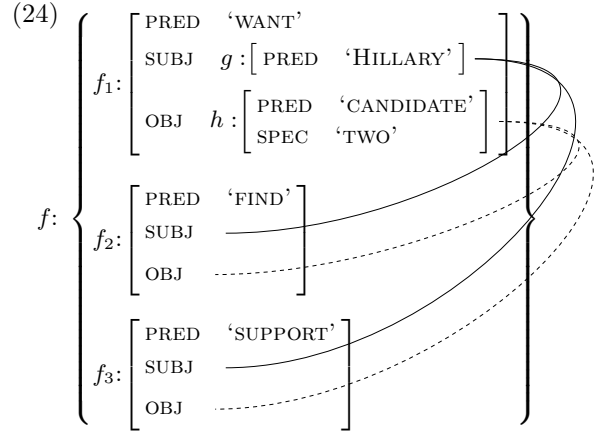
\enumsentence{\label{int-struct} \singlespace \evnup{
\hspace*{-2em}\fd{$f$:
\fdset{\fdstack
{$f_{1}$:\fdand{\feat{\sc pred}{`{\sc want}'}
       \feat{\sc subj}{$g:$\node{h}{\fdand{\feat{\sc pred}{`{\sc Hillary}'}}}}
       \feat{\sc obj}{$h:$\node{d}{\fdand{\feat{\sc pred}{`{\sc candidate}'}
                     \feat{\sc spec}{`{\sc two}'}}}}}}
{\fdstack{$f_{2}$:\fdand{\feat{\sc pred}{`{\sc find}'}
       \feat{\sc subj}{\node{i}{\strut}}
       \feat{\sc obj}{\node{e}{\strut}}}}
{$f_{3}$:\fdand{\feat{\sc pred}{`{\sc support}'}
       \feat{\sc subj}{\node{j}{\strut}}
       \feat{\sc obj}{\node{f}{\strut}}}}}}}
\nodecurve[r]{h}[r]{i}{9em}
\nodecurve[r]{h}[r]{j}{10em}
{\makedash{.4ex}
\nodecurve[r]{d}[r]{e}{9em}
\nodecurve[r]{d}[r]{f}{10em}}}}
The meaning constructors for the lexical items are given in Figure
\ref{fig:1}.  Recall that a second meaning constructor \BF{and2} is
introduced by {\it and} in order to handle cases where there are more
than two conjuncts; this contribution will be used once in the
derivation of the meaning for sentence (\ref{mixed-int-ent}). The
following \mbox{R-relations} result from the \mbox{f-structural}
relationships:

\[
\begin{array}{@{\strut}ll@{\strut}}
(i) & \Rrel{f}{conj}{f_1} \\
(ii) & \Rrel{f}{conj}{f_2} \\
(iii) & \Rrel{f}{conj}{f_3} \\
(iv) & \Rrel{f_1}{subj}{g} \\
(v) & \Rrel{f_2}{subj}{g} \\
(vi) & \Rrel{f_3}{subj}{g} \\
(vii) & \Rrel{f_1}{obj}{h} \\
(viii) & \Rrel{f_2}{obj}{h} \\
(ix) & \Rrel{f_3}{obj}{h} \\
\end{array}
\]
Following the analysis given in Dalrymple et al.\
\shortcite{DLPS:Intensional}, the lexical entry for {\it want} takes a
quantified NP as an argument.  This requires that the quantified NP
meaning be duplicated, since otherwise no readings result.  We provide
a special rule for duplicating quantified NPs when necessary:
\pagebreak
\enumsentence{$\BF{QNP Duplication}\colon$ \\[1ex]
$\begin{array}{@{\strut}l@{\strut}l@{\strut}} !(\All{F,Q}\\ & [\All{H,S}
(\All{ x} F_{\sigma}\means x \linimp H \means S(x)) \\ & \qquad
\linimp H \means Q(S)] \\
\linimp
& [\begin{array}[t]{@{\strut}l@{\strut}l@{\strut}}
&[\All{H,S} (\All{ x} F_{\sigma}\means x \linimp  H \means S(x)) \\
&\qquad \linimp H \means Q(S)] \\
\otimes & \hspace{.02in} [\All{H,S} (\All{ x} F_{\sigma}\means x \linimp  H
\means S(x)) \\
& \qquad \linimp H \means Q(S)]\hspace{.04in}\/]\/)\end{array} \\
\end{array}$}
In the interest of space, again we only show a few steps of the derivation.
Combining the meanings for {\it Hillary, found, supported,} and {\it and},
Axiom I, and \mbox{R-relations} (ii), (iii),
(v), (vi), (viii), and (ix), we can derive:
\[\begin{array}[t]{l}
h_{\sigma} \means x \linimp
f_{\sigma} \means and(\begin{array}[t]{@{\strut}l@{\strut}}
\IT{found}(\IT{Hillary}, x),\\ \IT{supported}\/(\IT{Hillary}, x)))
\end{array}\end{array}
\]
We duplicate the meaning of {\it two candidates} using QNP
Duplication, and combine one
copy with the foregoing formula to yield:
\[\begin{array}[t]{@{\strut}l@{\strut}}
f_{\sigma} \means two\/(z, \begin{array}[t]{@{\strut}l@{\strut}}candidate(z),\\
and(\begin{array}[t]{@{\strut}l@{\strut}}
\IT{found}(\IT{Hillary}, z),\\ \IT{supported}\/(\IT{Hillary}, z)))
\end{array}\end{array}\end{array}
\]
We then combine the other meaning of {\it two candidates} with the
meanings of {\it Hillary} and
{\it wanted}, and using Axiom I and \mbox{R-relations} (i), (iv), and
(vii) we obtain:
\[\begin{array}[t]{@{\strut}l@{\strut}}
\attr{f}{\sc conj}_{\sigma} \means\\
\qquad wanted\/(\begin{array}[t]{@{\strut}l@{\strut}}Hillary,\\
\hat{\ }\lambda{Q}.two\/(z,candidate(z),[\check{\ }{Q}](z)))
\end{array}\end{array}
\]
Finally, using {\it and2} with the two foregoing formulae, we deduce
the desired result:

\[\begin{array}[t]{@{\strut}l@{\strut}}
f_{\sigma} \means
and(wanted\/(\begin{array}[t]{@{\strut}l@{\strut}}\IT{Hillary}, \\
\hat{\ }\lambda {Q}.two\/(x, candidate(x), [\check{\ }{Q}](x))),\end{array} \\
\hspace*{.65in} two\/(z,\begin{array}[t]{@{\strut}l@{\strut}}
candidate(z),\\
and(\begin{array}[t]{@{\strut}l@{\strut}}
\IT{found}(\IT{Hillary}, z),\\ \IT{supported}\/(\IT{Hillary}, z))))
\end{array}\end{array}\end{array}\]
We can now specify a Partee and Rooth style processing strategy, which
is to prefer readings which require the least use of QNP duplication.
This strategy predicts the readings generated for the examples in
Section~\ref{sec:examples}.  It also predicts the desired reading for
sentence (\ref{mixed-int-ent}), since that reading requires two
quantifiers.  While the reading generated by Partee and Rooth is
derivable, it requires three quantifiers and thus uses QNP duplication
twice, which is less preferred than the reading requiring two
quantifiers which uses QNP~duplication once.  Also, it allows some
flexibility in cases where pragmatics strongly suggests that
quantifiers are copied and distributed for multiple extensional verbs;
unlike the Partee and Rooth account, this would apply equally to the
case where there are also intensional verbs and the case where there
are not.  Finally, our account readily applies to cases of intensional
verbs without coordination as in example (\ref{mixed-no-coord}), since
it applies more generally to cases of resource sharing.

\section{Conclusions and Future Work}
\label{sec:conclusion}

We have given an account of resource sharing in the syntax/semantics
interface of LFG.  The multiple use of semantic contributions results
from viewing dependencies in f-structures as resources; in this way
the one-to-one correspondence between \mbox{f-structure} relations and
meanings is maintained.  The resulting account does not suffer from
overgeneration inherent in other approaches, and applies equally to
cases of resource sharing that do not involve coordination.
Furthermore, it lends itself readily to an extension for the
intensional verb case that has advantages over the widely-assumed
account of Partee and Rooth
\shortcite{ParteeRooth:Conjunction}.

Here we have separated the issue of arriving at the appropriate
\mbox{f-structure} in the syntax from the issue of deriving the correct
semantics from the \mbox{f-structure}.  We have argued that this is the
correct distinction to make, and have given a treatment of the second
issue.  A treatment of the first issue will be articulated in a future
forum.

\section*{Acknowledgements}

We would like to thank Sam Bayer, John Maxwell, Fernando Pereira, Dick
Oehrle, Stuart Shieber, and especially Ron Kaplan for helpful
discussion and comments.  The first author was supported in part by
National Science Foundation Grant IRI-9009018, National Science
Foundation Grant IRI-9350192, and a grant from the Xerox Corporation.

\end{document}